Published in IEEE Access, Vol. 14, pp. 97418–97431, in 2026 (Digital Object Identifier 10.1109/ACCESS.2026.3707284)

# LRConv–NeRV: Low-Rank Convolution for Efficient Neural Video Compression

**Tamer Shanableh, Senior IEEE member**
American University of Sharjah, UAE,
Department of Computer Science and Engineering

## ABSTRACT

Neural Representations for Videos (NeRV) encode entire video sequences within neural network parameters, offering an alternative paradigm to conventional video codecs. However, the convolutional decoder of NeRV remains computationally expensive and memory intensive, limiting its deployment in resource-constrained environments. This paper proposes LRConv–NeRV, an efficient NeRV variant that replaces selected dense 3×3 convolutional layers with structured low-rank separable convolutions, trained end-to-end within the decoder architecture. By progressively applying low-rank factorization from the largest to earlier decoder stages, LRConv–NeRV enables controllable trade-offs between reconstruction quality and efficiency. Extensive experiments demonstrate that applying LRConv only to the final decoder stage reduces decoder complexity by 68%, from 201.9 to 64.9 GFLOPs, and model size by 9.3%, while incurring negligible quality loss and achieving approximately 9.2% bitrate reduction. Under INT8 post-training quantization, LRConv–NeRV preserves reconstruction quality close to the dense NeRV baseline, whereas more aggressive factorization of early decoder stages leads to disproportionate quality degradation. Compared to existing work under layer-aligned settings, LRConv–NeRV achieves a more favorable efficiency versus quality trade-off, offering substantial GFLOPs and parameter reductions while maintaining higher PSNR/MS-SSIM and improved temporal stability. Temporal flicker analysis using LPIPS further shows that the proposed solution preserves temporal coherence close to the NeRV baseline, results establish LRConv–NeRV as a potential architectural alternative for efficient neural video decoding under low-precision and resource-constrained settings.

## I. INTRODUCTION

Recent advances in implicit neural video representations, such as NeRV [1], have demonstrated that entire video sequences can be encoded within the parameters of a neural network and decoded directly from a frame index. Such a representation an alternative to video codecs that depend on digital signal processing and information theory [2] and [3] with overfitted deep learning models.

While implicit neural video representations offers attractive compression properties and simplifies the video coding pipeline, the decoder network remains computationally demanding due to the presence of high-capacity convolutional blocks operating at high spatial resolutions. These layers dominate both inference cost and model size, limiting the practical deployment of NeRV-based decoders on resource-constrained platforms.

To address this limitation, we propose LRConv–NeRV, a low-rank convolutional formulation for efficient NeRV decoding. The key idea is to replace selected dense spatial convolutions in the NeRV decoder with structured low-rank separable convolutions that approximate the original convolutional operators with significantly fewer parameters and floating-point operations [4]. Unlike post-hoc low-rank decomposition methods that factorize pretrained kernels, LRConv–NeRV enforces the low-rank structure architecturally and trains it end-to-end, ensuring that the decoder adapts naturally to the reduced representational capacity.

LRConv–NeRV differs fundamentally from prior NeRV efficiency variants such as Binary-NeRV [5] and Ghost-NeRV [6]. Binary-NeRV achieves extreme compression by discretizing convolutional weights to binary values, which substantially reduces computation but can severely restrict representational capacity and introduce noticeable reconstruction artifacts. Ghost-NeRV, on the other hand, reduces FLOPs by generating feature maps through inexpensive linear transformations from a small set of intrinsic features, relying on redundancy among output channels. In contrast, LRConv–NeRV reduces complexity by constraining the intrinsic rank of the convolutional transformation itself, preserving continuous-valued weights and full channel mixing through a principled factorization of the convolution kernel.

Through extensive rate–distortion and complexity analysis, we show that applying LRConv to the largest decoder stages yields a favorable trade-off between efficiency and reconstruction quality, achieving substantial reductions in GFLOPs and model size with minimal loss in PSNR and MS-SSIM. Moreover, LRConv–NeRV remains robust under low-precision inference and preserves temporal coherence, making it potentially a practical architectural mechanism for deploying implicit neural video representations in resource-constrained decoding environments.

## II. Literature Review

Recent advances in implicit neural video representations, such as NeRV [1], have demonstrated that entire video sequences can be encoded within the parameters of a neural network and decoded directly from a frame index. Such a representation an alternative to video codecs that depend on digital signal processing and information theory [2] and [3] with overfitted deep learning models.

While implicit neural video representations offer attractive compression properties and simplifies the video coding pipeline, the decoder network remains computationally demanding due to the presence of high-capacity convolutional blocks operating at high spatial resolutions. These layers dominate both inference cost and model size, limiting the practical deployment of NeRV-based decoders on resource-constrained platforms.

To address this limitation, we propose LRConv–NeRV, a low-rank convolutional formulation for efficient NeRV decoding. The key idea is to replace selected dense spatial convolutions in the NeRV decoder with structured low-rank separable convolutions that approximate the original convolutional operators with significantly fewer parameters and floating-point operations [4]. Unlike post-hoc low-rank decomposition methods that factorize pretrained kernels, LRConv–NeRV enforces the low-rank structure architecturally and trains it end-to-end, ensuring that the decoder adapts naturally to the reduced representational capacity.

LRConv–NeRV differs fundamentally from prior NeRV efficiency variants such as Binary-NeRV [5] and Ghost-NeRV [6]. Binary-NeRV achieves extreme compression by discretizing convolutional weights to binary values, which substantially reduces computation but can severely restrict representational capacity and introduce noticeable reconstruction artifacts. Ghost-NeRV, on the other hand, reduces FLOPs by generating feature maps through inexpensive linear transformations from a small set of intrinsic features, relying on redundancy among output channels. In contrast, LRConv–NeRV reduces complexity by constraining the intrinsic rank of the convolutional transformation itself, preserving continuous-valued weights and full channel mixing through a principled factorization of the convolution kernel.

## III. Proposed Low-Rank Convolution for Efficient NeRV Decoding

The Neural Representation for Videos (NeRV) decoder employs a sequence of convolutional blocks to progressively upsample a compact latent embedding into full-resolution video frames. While effective, these convolutional layers dominate the computational cost and parameter footprint of the model, particularly in the later decoder stages where channel dimensionality is high prior to pixel-shuffle upsampling.

To reduce decoding complexity while preserving reconstruction quality, we introduce Low-Rank Convolution (LRConv) into the NeRV decoder. Our approach replaces selected dense spatial convolutions with a structured low-rank factorization that is trained end-to-end as part of the network. This avoids post-hoc kernel decomposition like SVD and enables substantial reductions in parameters and FLOPs without additional training stages.

The overall architecture of the proposed solution is depicted in Figure 1. Note that NeRV has five decoding stages or five convolution layers for reconstructing the full resolution output image. To illustrate the proposed architecture of Figure 1, consider a standard convolutional layer in NeRV with kernel size $k \times k$, $C_{in}$ input channels, and $C_{out}$ output channels. Given an input feature map $\mathbf{X} \in \mathbb{R}^{C_{in} \times H \times W}$, the output feature map $\mathbf{Y}$ is computed as:

$$\mathbf{Y}_{c_o}(h, w) = \sum_{c_i=1}^{C_{in}} \sum_{u=1}^{k} \sum_{v=1}^{k} \mathbf{W}_{c_o, c_i, u, v} \, \mathbf{X}_{c_i}(h+u, w+v) \quad (1)$$

The parameter count and computational complexity scale as $O(k^2 C_{in} C_{out})$, which becomes the dominant cost in later NeRV decoding stages due to large channel dimensions.

In the proposed solution, a full convolution kernel can be reshaped into a matrix $\widetilde{\mathbf{W}} \in \mathbb{R}^{C_{out} \times (C_{in} k^2)}$. Instead of learning a dense kernel, LRConv constrains this matrix to be low-rank as follows:

$$\widetilde{\mathbf{W}} \approx \mathbf{AB} \quad (2)$$

$$\mathbf{A} \in \mathbb{R}^{C_{out} \times r}$$

$$\mathbf{B} \in \mathbb{R}^{r \times (C_{in} k^2)}$$

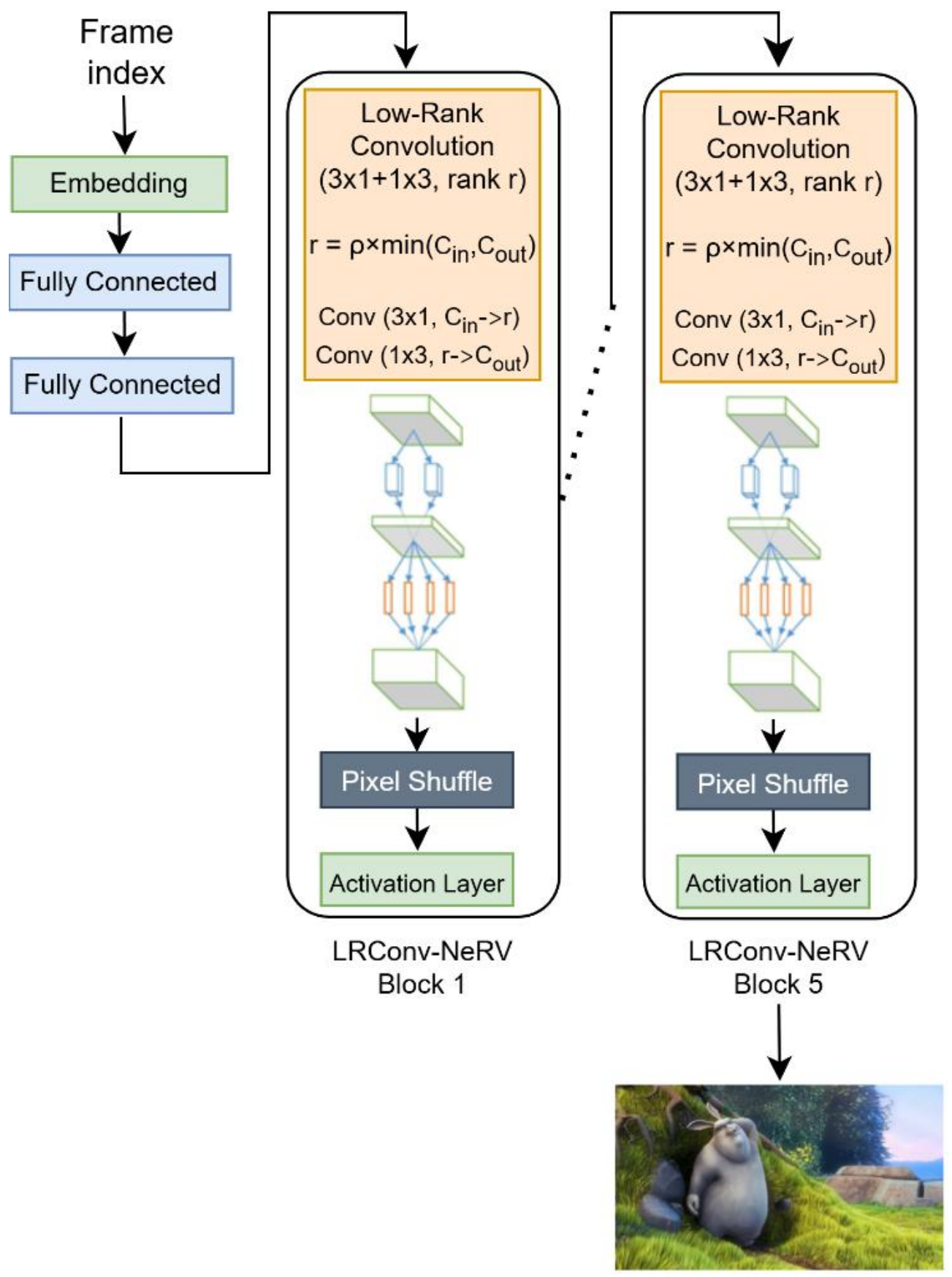

**Figure 1.** Overall architecture of the proposed LRConv-NeRV solution.

Where the rank $r$ is selected as a fraction of the channel dimension:

$$r = \lceil \rho \cdot \min(C_{in}, C_{out}) \rceil \quad (3)$$

where $\rho \in (0,1]$ controls the degree of factorization. All LRConv layers are trained end-to-end with the NeRV decoder. In this work we set the LRConv bottleneck ratio to $\rho = 0.25$, as it provides substantial reductions in the replaced convolutional cost while remaining large enough to preserve NeRV's reconstruction fidelity in the late decoder stage

Consequently, rather than computing this factorization post-training, we enforce the low-rank structure architecturally by decomposing the spatial convolution into two separable convolutions:

$$\mathrm{Conv}(k \times k, C_{in} \to C_{out}) \to \mathrm{Conv}(k \times 1, C_{in} \to r) \circ \mathrm{Conv}(1 \times k, r \to C_{out}) \quad (4)$$

Where the symbol "∘" denotes function composition which is a sequential application.

This factorization reduces the convolution to a composition of vertical and horizontal filtering with a reduced intermediate rank. Given an input feature map **X**, LRConv proceeds in two steps, projection, which is low-rank encoding, and reconstruction, which is channel expansion. In the projection we have:

$$\mathbf{Z} = \mathrm{Conv}_{k\times 1}\left(\mathbf{X}; \mathbf{W}^{(1)}\right) \quad (5)$$
$$\mathbf{W}^{(1)} \in \mathbb{R}^{r\times C_{in}\times k\times 1}$$
$$\mathbf{Z} \in \mathbb{R}^{r\times H\times W}$$

And in the reconstruction, we have:

$$\mathbf{Y} = \mathrm{Conv}_{1\times k}\left(\mathbf{Z}; \mathbf{W}^{(2)}\right) \quad (6)$$
$$\mathbf{W}^{(2)} \in \mathbb{R}^{C_{out}\times r\times 1\times k}$$

Where the superscripts on the weight matrices $\mathbf{W}^{(1)}$ and $\mathbf{W}^{(2)}$ denote the projection and reconstruction stages of the factorized convolution, respectively. The intermediate representation **Z** can be interpreted as a low-dimensional embedding of local spatial–channel patches, while the second convolution reconstructs the full output feature map. Consequently, the parameter count of LRConv is:

$$\mathrm{Params}_{\mathrm{LR}} = k\, r\, (C_{in} + C_{out}) \quad (7)$$

Compared to the dense convolution

$$\mathrm{Params}_{\mathrm{full}} = k^2 C_{in} C_{out} \quad (8)$$

For $r \ll \min(C_{in}, C_{out})$, LRConv yields substantial savings:

$$\frac{\mathrm{Params}_{\mathrm{LR}}}{\mathrm{Params}_{\mathrm{full}}} = \frac{r(C_{in}+C_{out})}{k\, C_{in}\, C_{out}} \quad (9)$$

In NeRV, these savings are impactful because LRConv is applied before pixel-shuffle upsampling, where channel counts are high and dominate the decoder's computational cost.

In this proposed solution, LRConv is applied selectively to a subset of NeRV decoder stages. Let the decoder consist of $S$stages indexed from low to high resolution. We define $\mathcal{S}_{\mathrm{LR}} \subseteq \{0, \ldots, S-1\}$ as the set of stages where dense convolutions are replaced by LRConv. This stage-wise design enables fine-grained control over the efficiency versus quality trade-off and forms the basis for the experimental configurations evaluated in the experimental results section.

Lastly, to illustrate the effect of low-rank factorization in the NeRV decoder, consider the last decoding stage of the five decoding stages in NeRV, L4, where a standard 3×3 convolution maps $C_{in} = 96$ input channels to $C_{out} = 384$ output channels. Using the rank selection rule $r = \lceil \rho \cdot \min(C_{in}, C_{out}) \rceil$with $\rho = 0.25$, the dense convolution is replaced by two separable convolutions: a 3×1 projection $96 \to 24$ followed by a 1×3 expansion $24 \to 384$. This reduces the number of convolutional parameters from $9 \cdot 96 \cdot 384$ in the dense case to $3 \cdot 24 \cdot 96 + 3 \cdot 24 \cdot 384$, corresponding to an approximately 9.6× reduction in weight parameters for this layer. A similar proportional reduction applies to the number of multiply–accumulate operations. Thus, enabling substantial savings in GFLOPs and model size with minimal impact on reconstruction quality when applied to later decoder stages.

### IV. Comparison with Binary-NeRV and Ghost-NeRV

Low-rank convolution provides a complementary efficiency mechanism to Binary-NeRV [5] and Ghost-NeRV [6], differing fundamentally in how computational complexity is reduced and how reconstruction fidelity is preserved within the NeRV decoding pipeline.

Binary-NeRV achieves substantial compression by binarizing selected convolutional layers, replacing floating-point weights with binary values and introducing scaling factors to partially recover magnitude information. While this leads to significant reductions in model size and arithmetic cost, it imposes a hard quantization constraint on decoder weights. This severely restricts the representational capacity of the affected layers and results in a sharp accuracy versus efficiency trade-off. Consequently, Binary-NeRV typically results in visual degradations and temporal flicker. In contrast, low-rank convolution operates entirely in the real-valued domain and reduces complexity by constraining the intrinsic rank of the convolutional operator rather than its numerical precision. This preserves continuous optimization and avoids the discrete optimization challenges inherent to binarization, leading to more graceful degradation in reconstruction quality under increasing compression.

On the other hand, Ghost-NeRV reduces computation by decomposing convolutional layers into a small set of intrinsic feature maps followed by inexpensive linear transformations that generate additional ghost features. Although effective in lowering FLOPs, this approach relies on the assumption that many output channels are redundant and can be synthesized from a limited basis. In the context of NeRV decoding, where convolutional layers are responsible for reconstructing fine-grained spatial detail from implicit representations, such heuristic feature generation can limit expressiveness of the decoder's convolutional mapping from latent features to pixel-space details.

By contrast, low-rank convolution preserves full channel mixing through a structured factorization of the convolution kernel itself, yielding a reduction in complexity while maintaining a global linear mapping between input and output channels.

From a modeling perspective, Binary-NeRV reduces complexity by constraining weight precision, Ghost-NeRV reduces complexity by approximating feature generation, whereas LRConv reduces complexity by limiting the intrinsic dimensionality of the convolutional transformation. These methods therefore target orthogonal aspects of efficiency.

### V. Experimental Results

In this section, we evaluate the proposed LRConv-NeRV in terms of rate–distortion (RD) performance, temporal flicker, model size, and computational complexity measured in GFLOPs. The evaluation is conducted using progressive replacement of dense convolution layers with low-rank convolutions, enabling a systematic analysis of the resulting trade-offs between reconstruction quality and efficiency. Experiments are performed on the Bunny sequence and three additional sequences from the UVG dataset [29]. Following the experimental protocol of Binary-NeRV and Ghost-NeRV, all sequences are resized to 720x1280 to match the Bunny resolution and contain 132 frames each. Representative frames from each sequence are shown in Figure 2.

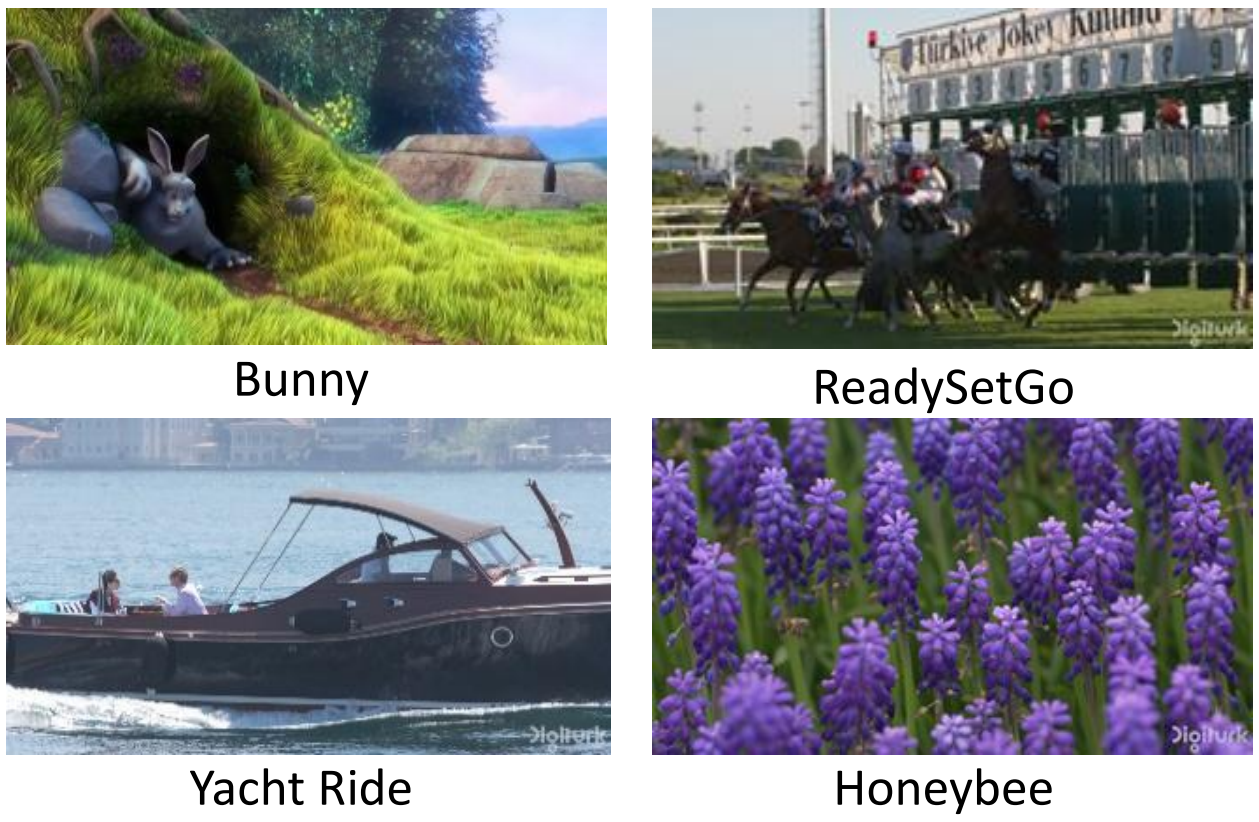

**Figure 2**. video sequences used in the evaluation of the proposed LRConv-NeRV solution.

#### *(a) Rate–Distortion (RD) Results*

We evaluate the rate–distortion (RD) performance of the proposed LRConv–NeRV under different layer replacement configurations and rank settings. RD curves are reported in terms of PSNR versus bitrate (bpp), with the dense NeRV serving as the baseline. Results are grouped by the set of decoder stages where 3×3 convolutions are replaced with low-rank separable convolutions, applied progressively from the largest to the smallest convolution layers (i.e., from later to earlier decoder stages).

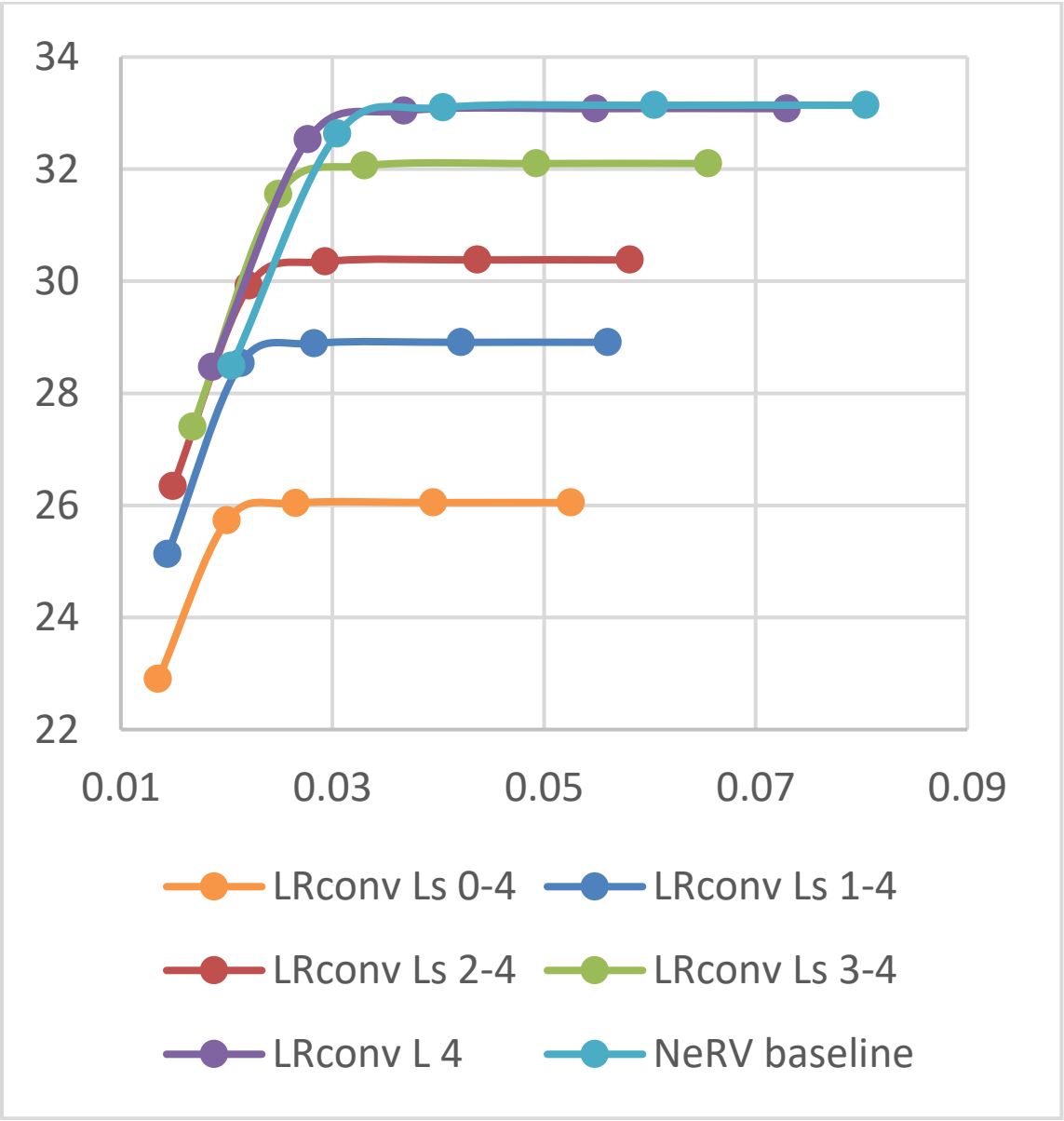

(a) Bunny Sequence

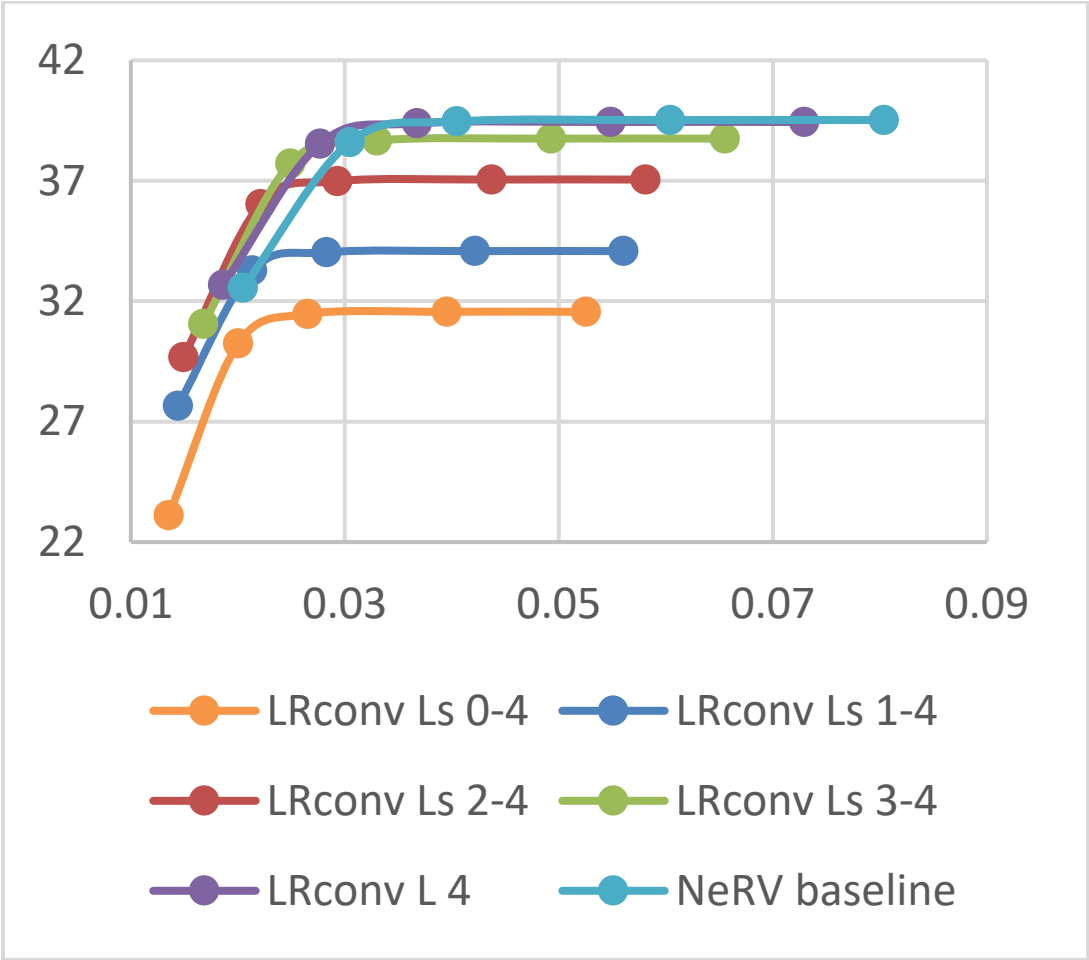

(b) HoneyBee Sequence

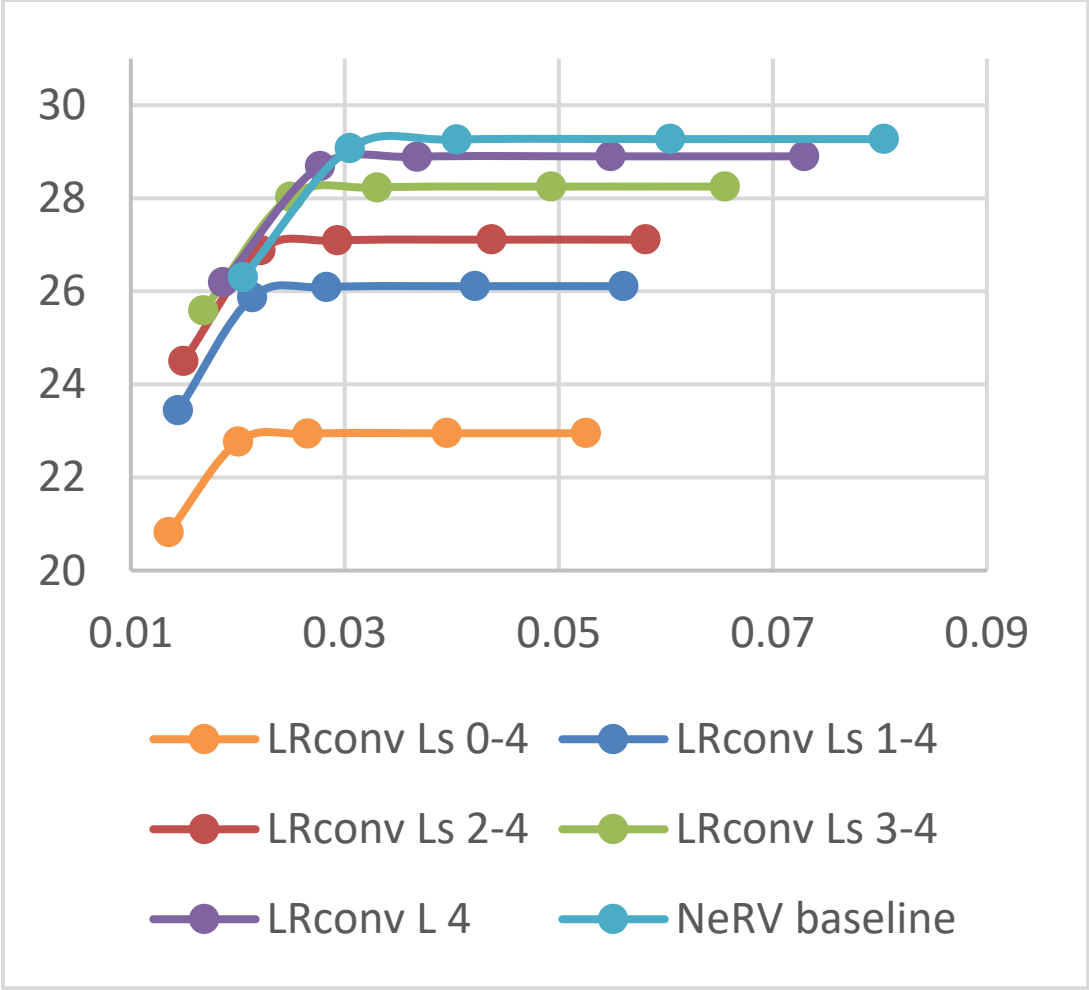

(c) ReadySetGo Sequence

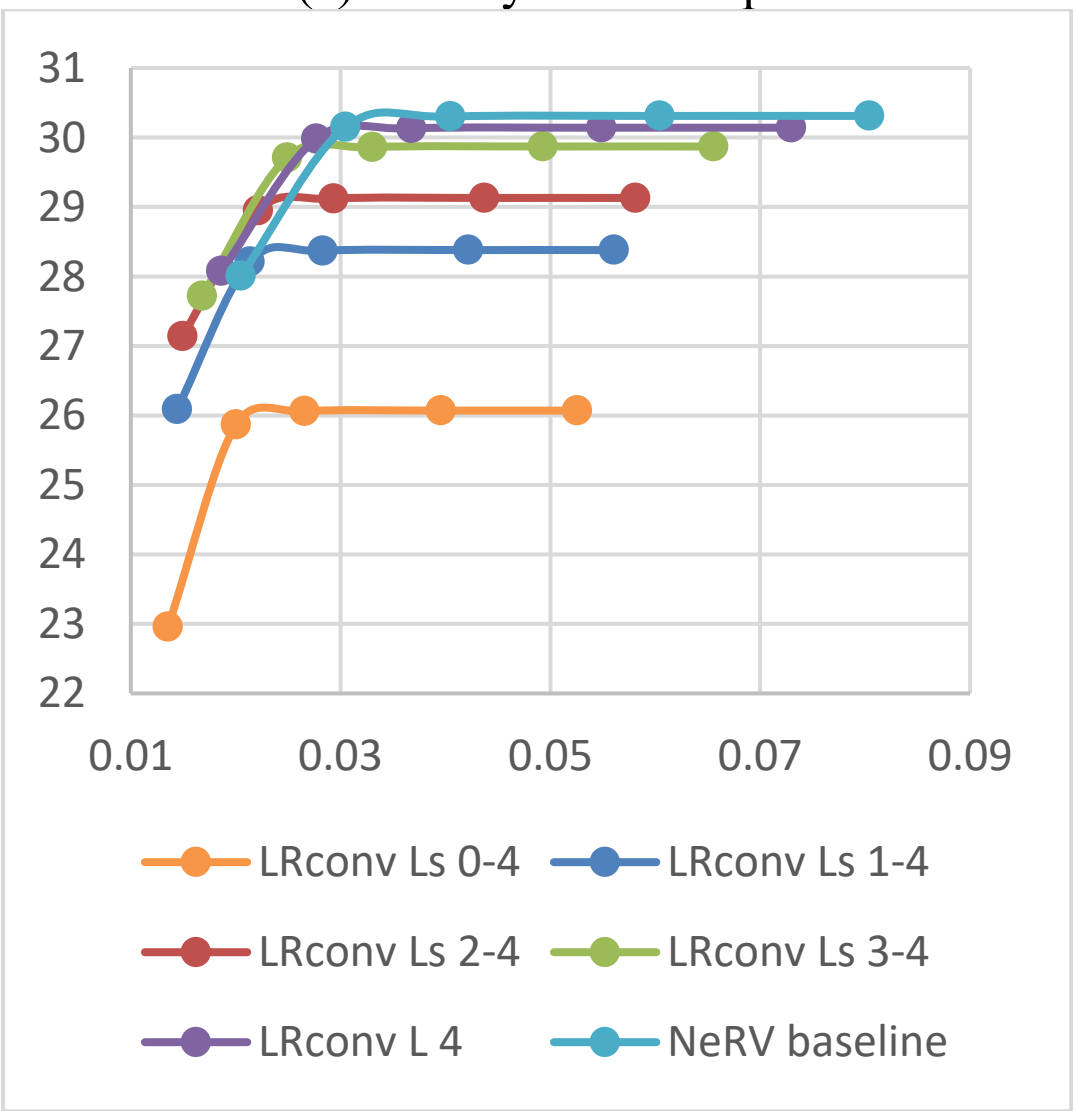

(d) YachtRide Sequence

**Figure 3.** RD results of the proposed LRConv–NeRV with progressive replacement of convolution layers from the largest to the smallest stages.

Across all sequences, LRConv–NeRV exhibits a consistent RD trade-off: replacing later decoder stages yields substantial bitrate and complexity reductions with limited PSNR degradation, while progressively extending LRConv to earlier stages increases distortion for marginal additional rate savings. This behavior aligns with the structure of NeRV, where later stages operate at higher spatial resolutions and dominate the decoder's computational and representational capacity.

Applying LRConv only at the final decoder stage (L4) results in a small PSNR drop relative to dense NeRV at comparable bitrates, while already yielding a noticeable reduction in rate. Extending LRConv to stages L3–L4 further improves compression efficiency, with modest additional distortion. In contrast, configurations that include LRConv in early stages (e.g., L0–L4) exhibit a more pronounced quality loss for comparatively smaller gains in rate reduction. This indicates that early-stage factorization disrupts the global structure encoded in lower-resolution feature maps, which propagates error through subsequent upsampling stages.

The RD behavior is consistent across different content types, though texture-rich sequences (e.g., fast motion or fine detail) exhibit slightly larger quality drops under aggressive LRConv configurations. This suggests that the low-rank factorization more strongly impacts the modeling of high-frequency content, which becomes more prominent in later decoder stages operating at full resolution.

Overall, the RD results demonstrate that LRConv–NeRV offers a controllable trade-off between bitrate and reconstruction quality. Replacing convolutions in later decoder stages achieves the most favorable RD operating points, while aggressive factorization of early stages should be avoided due to disproportionate quality loss. These findings motivate stage-wise LRConv as a practical mechanism for adapting NeRV to resource-constrained decoding environments.

More specifically, Table 2 reports the average differences between the RD curves of the proposed solution applied to the last decoder convolution layer and the dense NeRV baseline.

**Table 2.** Average differences between the RD curves of LRConv applied to the final convolution layer and the NeRV baseline.

| | **Δ PSNR** | **Δ ms-ssim** | **% BPP reduction** | **% Params reduction** |
|---|---|---|---|---|
| **Bunny** | -0.06 | -9.20E-04 | 9.19 | 9.28 |
| **HoneyBee** | -0.038 | -4.00E-05 | 9.19 | 9.28 |
| **ReadySetGo** | -0.32 | -3.98E-03 | 9.19 | 9.28 |
| **YachtRide** | -0.122 | -2.06E-03 | 9.19 | 9.28 |

The results indicate that the average PSNR and MS-SSIM differences are negligible. In particular, PSNR differences below 0.5 dB are widely regarded as subjectively imperceptible, which is corroborated by the small MS-SSIM gaps observed across all sequences. Importantly, this near-baseline reconstruction quality is achieved alongside consistent reductions of approximately 9.2% in bitrate and 9.3% in model parameters, confirming the effectiveness of LRConv applied at later decoder stages as a strong RD operating point.

### *B. Complexity Analysis*

In this subsection, we analyze the computational complexity of the proposed LRConv–NeRV by progressively replacing dense 3×3 convolutions with low-rank separable convolutions from the largest to the smallest decoder stages. We report the decoder complexity in terms of GFLOPs (computed as 2×MACs) and examine the associated trade-offs in PSNR and model size as a function of computational cost. This analysis provides a quantitative view of how architectural factorization impacts both efficiency and reconstruction quality.

Table 3 reports the GFLOPs of LRConv–NeRV and compares them against two variants of Ghost-NeRV [6] and Binary-NeRV [5] under equivalent architectural settings. The layer indices indicate the set of decoder stages to which the modification is applied, progressing from the largest or highest-resolution stage toward earlier stages.

**Table 3**. GFLOPs of the proposed LRConv-NeRV versus existing work (The GLOPs of the NeRV baseline is 201.9)

| Convolution Layers | LRConv NeRV | Ghost NeRV[6] | GhostV2 NeRV[6] | Binary NeRV[5] |
|---|---|---|---|---|
| **4 (Largest)** | 64.92 | 126.24 | 142.96 | 72.60 |
| **4–3** | 30.68 | 107.32 | 128.36 | 34.60 |
| **4–2** | 22.12 | 102.60 | 124.71 | 25.20 |
| **4–1** | 21.54 | 102.29 | 124.47 | 24.50 |
| **4–0 (All)** | 21.50 | 102.26 | 124.46 | 24.50 |

LRConv–NeRV provides consistent and substantial GFLOP reductions with moderate architectural modification. Replacing only the largest convolution layer reduces complexity from the dense NeRV baseline (≈201.9 GFLOPs) to 64.9 GFLOPs, corresponding to a reduction of nearly 68% in decoder cost while preserving most of the reconstruction quality, as shown in the RD analysis.

Progressively extending LRConv to earlier layers further reduces GFLOPs to approximately 22.12 GFLOPs, after which additional replacements yield only marginal computational savings. This saturation effect reflects the diminishing contribution of early decoder stages to the overall computational burden compared to high-resolution layers.

Compared to Ghost-NeRV and GhostV2-NeRV, LRConv–NeRV consistently achieves significantly lower GFLOPs for the same layer configurations, indicating more efficient structured factorization of convolutional operators. While Binary-NeRV achieves lower computational cost in some configurations due to aggressive binarization, this reduction comes at the expense of notable spatial quality degradation and increased temporal flicker, as demonstrated in the RD and temporal stability results.From a deployment perspective, applying LRConv to the largest one decoder stage (L4) provides a strong operating point, achieving significant GFLOPs reduction with minimal RD degradation. More aggressive configurations primarily benefit scenarios with extreme compute constraints, where additional quality loss is acceptable.

Figure 4 reports the PSNR of the reconstructed videos under INT8 post-training quantization for LRConv–NeRV applied progressively from the largest decoder convolution layer (L4) toward earlier layers (L3–L4, L2–L4, L1–L4, and L0–L4). The results demonstrate that LRConv–NeRV maintains strong reconstruction quality under low-precision inference when factorization is limited to the largest decoder stages.

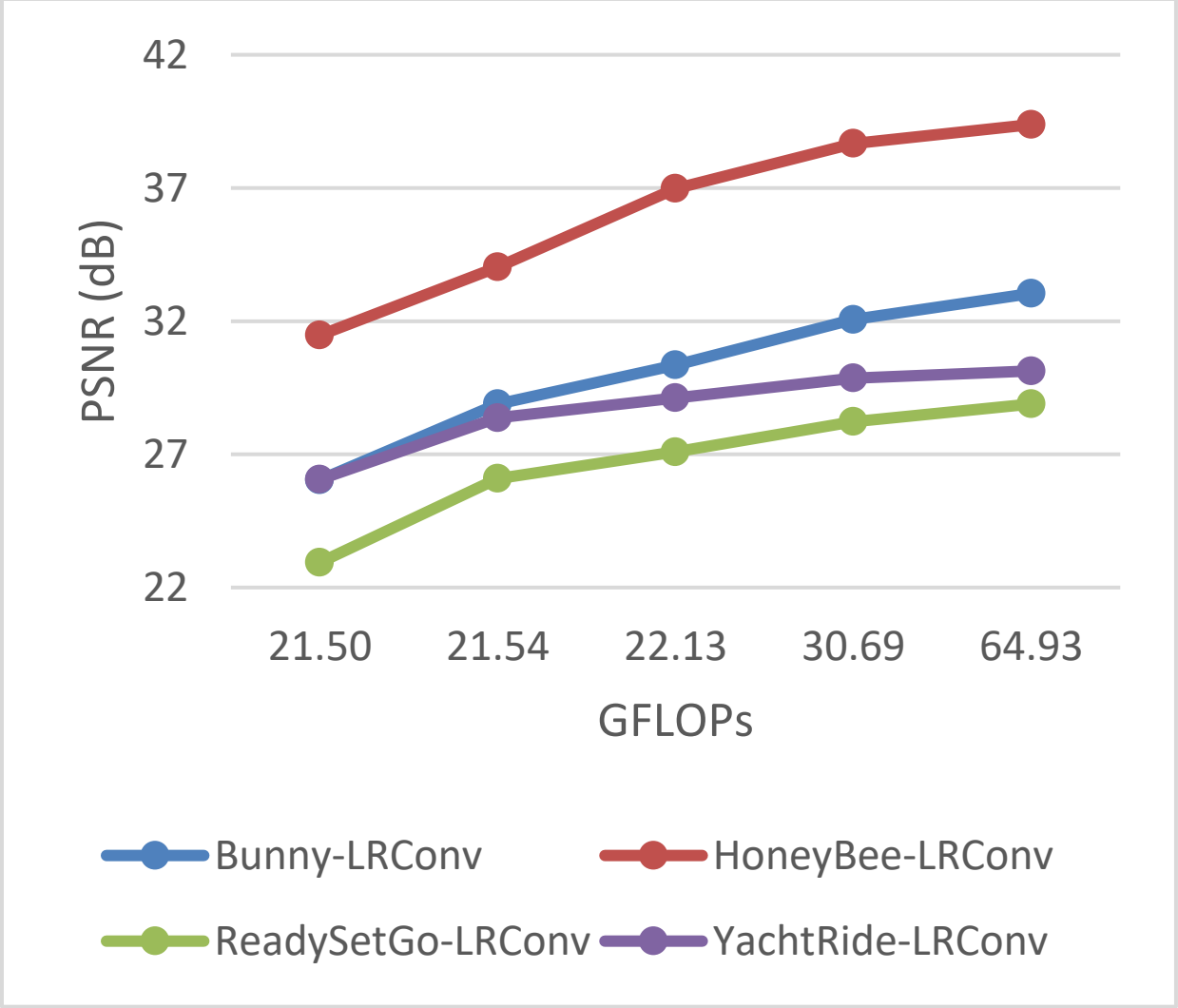

**Figure 4.** the PSNR of the reconstructed videos with a quantization accuracy of 8 bits when applying the proposed LRConv-NeRV to convolution layer 4.

When LRConv is applied only to the largest convolution layer (L4), the PSNR remains nearly identical to the dense NeRV baseline across all sequences, despite a substantial reduction in computational complexity (from 201.88 GFLOPs to 64.93 GFLOPs). The PSNR gap relative to NeRV is minimal for Bunny (−0.06 dB), HoneyBee (−0.08 dB), ReadySetGo (−0.37 dB), and YachtRide (−0.17 dB), indicating that architectural factorization at the final decoding stage is highly robust to INT8 quantization.

As LRConv is progressively extended to earlier layers (L3–L4 and L2–L4), PSNR degrades more noticeably, even though the additional reductions in GFLOPs are comparatively modest. For example, extending LRConv from L4 to L3–L4 reduces complexity from 64.93 to 30.68 GFLOPs, but introduces additional quality loss across all sequences. More aggressive configurations (L1–L4 and L0–L4) achieve the lowest complexity, 21.5 GFLOPs, but incur substantial degradation in reconstruction quality (e.g., Bunny drops from 33.04 dB at L4 to 28.89 dB at L1–L4 and 26.04 dB at L0–L4), highlighting that early-stage factorization under INT8 precision significantly harms the fidelity of the decoded frames.

Overall, Figure 4 confirms that applying LRConv only to the largest decoder stage (L4) yields the most favorable quality–complexity trade-off under INT8 quantization, while progressively factorizing earlier stages leads to diminishing efficiency gains and disproportionate quality loss.

Figure 5 plots the number of model parameters as a function of GFLOPs for LRConv–NeRV under the same experimental setup as Figure 4 (INT8 quantization, LRConv applied progressively from layer L4 to L0–4). The results demonstrate a clear and monotonic relationship between architectural factorization and model compactness.

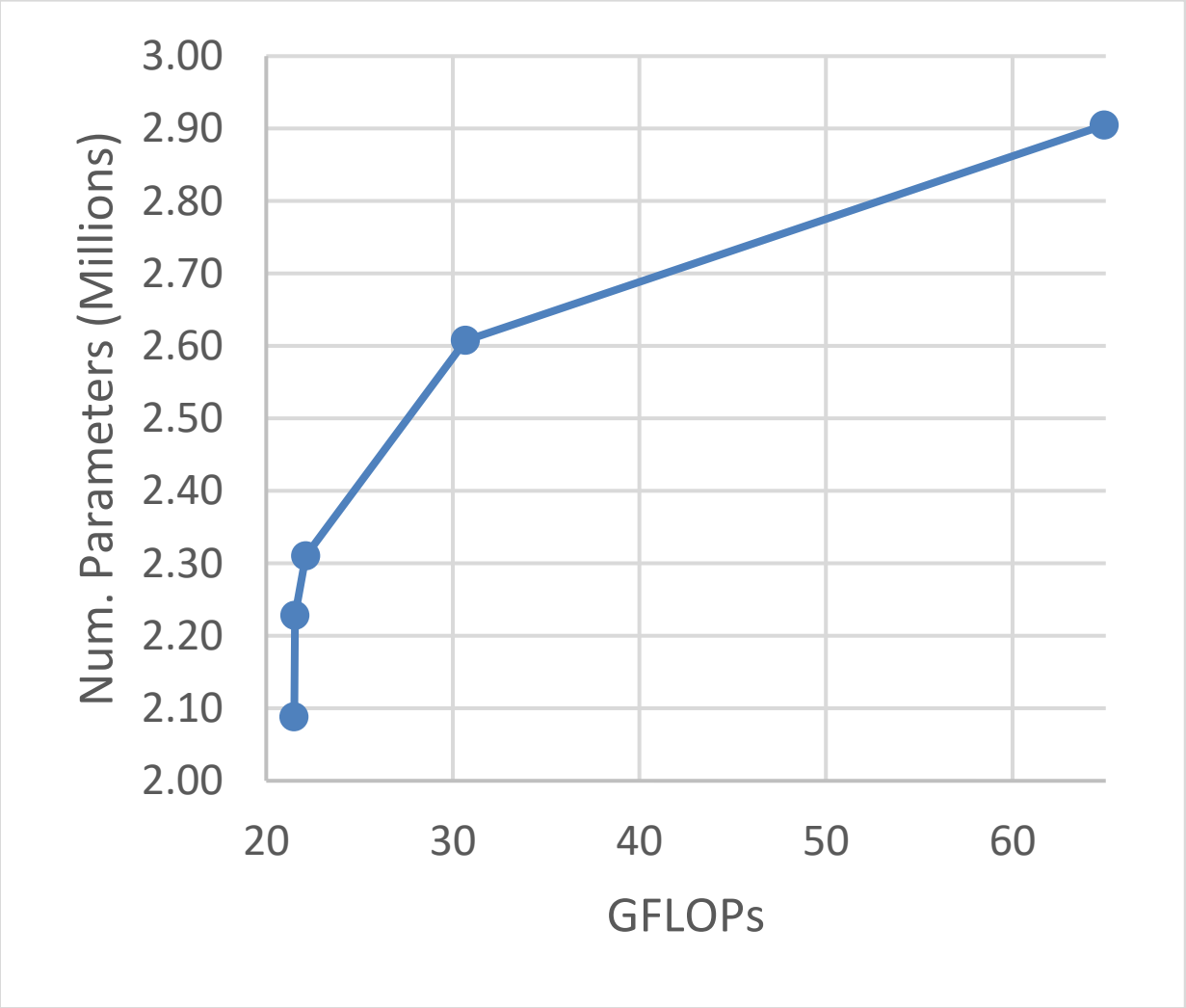

**Figure 5.** Plot the number of model parameters as a function of GFLOPs

Applying LRConv only to the largest decoder stage (L4) reduces the model size from the dense NeRV baseline (≈3.20M parameters) to 2.90M parameters while lowering the computational cost to 64.93 GFLOPs. Extending LRConv to stages L3–L4 further reduces the parameter count to 2.61M with a corresponding reduction in compute to 30.68 GFLOPs. More aggressive configurations (L2–L4 and L1–L4) continue to reduce both parameters and GFLOPs, reaching 2.31M and 2.23M parameters at approximately 22.12 GFLOPs, respectively. The most aggressive configuration (L0–L4) yields the smallest model with 2.09M parameters at 21.5 GFLOPs.

Overall, Figure 5 highlights that LRConv–NeRV offers a favorable efficiency–compactness trade-off, and that applying LRConv to the largest decoder stage(s) provides a practical operating point that substantially reduces both computational complexity and model size with minimal impact on reconstruction quality.

### *(c) Comparison with existing work*

Table 4 summarizes the per-frame computational cost (GFLOPs) of representative NeRV variants and the proposed LRConv–NeRV. The proposed method achieves 64.92 GFLOPs, providing a substantial reduction relative to the dense NeRV baseline (201.9 GFLOPs) and most prior variants (e.g., HiNeRV, DNeRV, HNeRV), while remaining competitive with Ghost-NeRV applied at L4. Binary–NeRV attains the second lowest computational cost of 71.4 GFLOPs. However, as shown in the RD and temporal flicker evaluations, this aggressive binarization leads to notable degradation in reconstruction quality and increased temporal instability. In contrast, LRConv–NeRV attains the lowest computational cost of 64.92 GFLOPs, delivering significant complexity reduction while preserving spatial fidelity and temporal coherence close to the dense NeRV baseline.

**Table 4**. Per-frame computational cost in GFLOPs in comparison to existing NeRV variants with 720x1280 resolution.

| **NeRV Variant** | **Gflops** |
|---|---|
| NeRV (Baseline) [1] | 201.9 |
| E-NeRV [20] | 208 |
| HNeRV [21] | 179 |
| FFNeRV [12] | 204 |
| SNeRV-B [9] | 363.6 |
| SNeRV-T [9] | 206.6 |
| HiNeRV [16] | 190 |
| DNeRV [10] | 181 |
| Binary-NeRV L4 [5] | 71.4 |
| Ghost-NeRV-v2 L4 [6] | 143 |
| Ghost-NeRV L4 [6] | 126.24 |
| Proposed sol. for L4 | 64.92 |

For comparison with more relevant existing solutions, we focus on applying LR convolution to largest convolution layer, which is L4. In the following we use INT8 quantization configuration to compare the PSNR, MS-SSIM, BPP and number of model parameters of the proposed solution against the 2 variants of Ghost-NeRV [6] and Binary-NeRV [5]. This comparison makes sense the mentioned NeRV variants apply Ghost convolution and Binary convolution on individual convolution layers similar to the proposed solution. Hence enabling a fair, layer-aligned comparison since all methods modify the same decoder stage.

As shown in Figure 6 (PSNR) and Figure 7 (MS-SSIM), LRConv–NeRV consistently preserves reconstruction quality closer to the dense NeRV baseline than Binary–NeRV [5], and is competitive with GhostConv variants [6] across all sequences. In particular, LRConv–NeRV exhibits a smaller quality gap to NeRV than Binary–NeRV on texture-rich content (e.g., ReadySetGo and YachtRide), indicating that low-rank factorization at L4 better retains high-frequency structure than aggressive binarization. Compared to GhostConv–v1/v2, LRConv–NeRV achieves comparable PSNR/MS-SSIM while offering a simpler factorized convolutional structure with predictable rank control.

Figure 8 (BPP) shows that LRConv–NeRV attains lower bitrate than dense NeRV and GhostConv variants [6] across sequences, while remaining higher than Binary–NeRV [5], which is expected due to the extreme compression of binary weights. Importantly, LRConv–NeRV provides a balanced

operating point where it achieves meaningful bitrate reductions without incurring the pronounced quality loss observed for Binary–NeRV, reflecting a more favorable rate–distortion trade-off at L4.

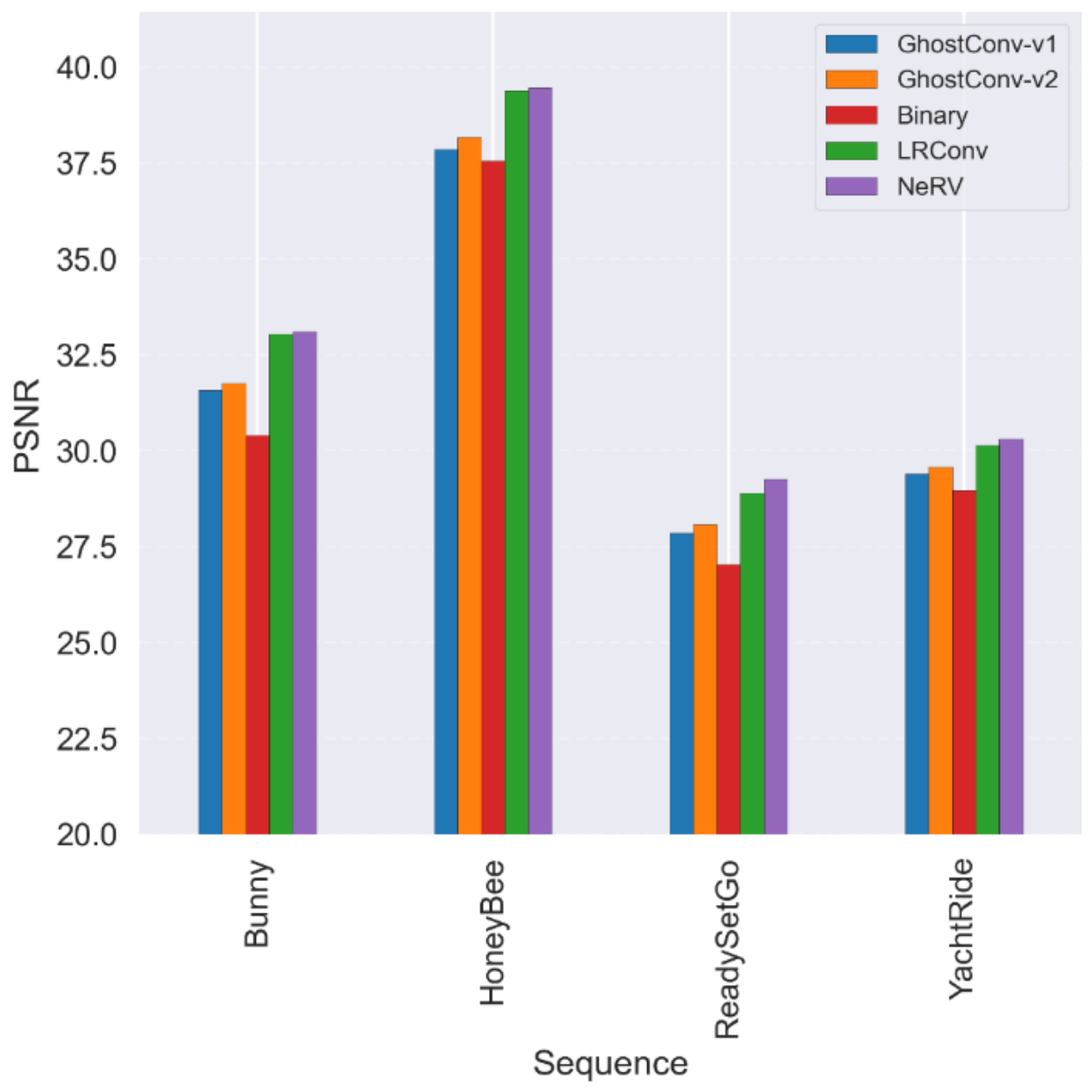

**Figure 6**. PSNR comparison with existing solutions applied to convolution layer 4 with INT8 quantization.

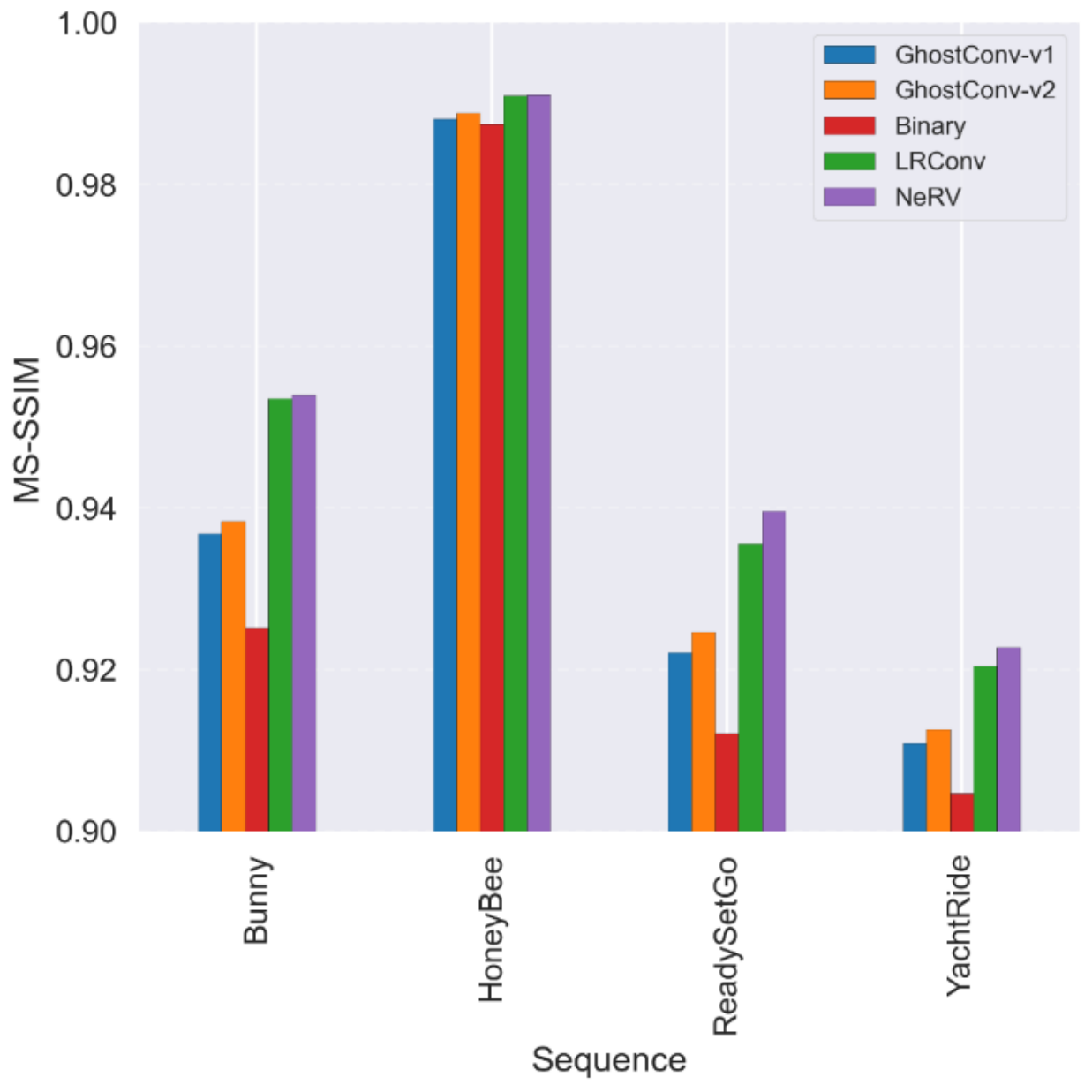

**Figure 7.** MS-SSIM comparison with existing solutions applied to convolution layer 4 with INT8 quantization.

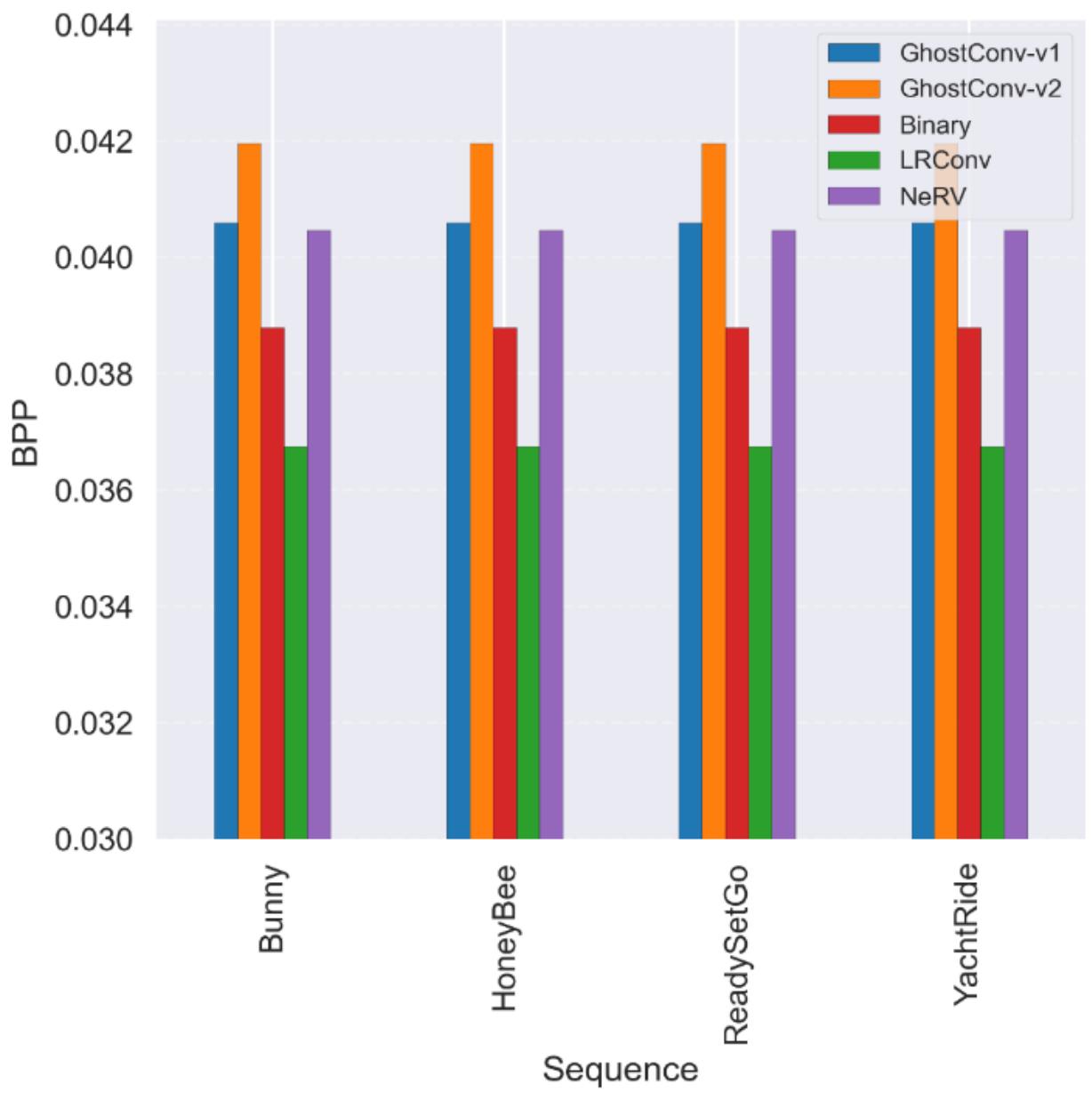

**Figure 8.** Bits-per-pixel comparison with existing solutions applied to convolution layer 4 with INT8 quantization.

In Figure 9 (number of parameters), LRConv–NeRV consistently yields a smaller model size than NeRV and GhostConv variants [6], while Binary–NeRV [5] achieves the smallest footprint. This confirms that LRConv reduces decoder capacity at the modified stage without the representational collapse associated with binarization. The parameter savings of LRConv–NeRV, combined with its near-baseline quality, highlight its practicality for deployment scenarios where both memory footprint and fidelity matter.

Overall, the comparison demonstrates that LRConv–NeRV occupies a favorable middle ground between GhostConv-based factorization and Binary–NeRV. This is so as the proposed solution offers substantial reductions in parameters and bitrate with significantly better reconstruction quality than Binary–NeRV, and quality comparable to GhostConv variants at a lower or similar model footprint. These results validate LRConv–NeRV as an effective architectural alternative for compressing NeRV decoders under low-precision inference constraints.

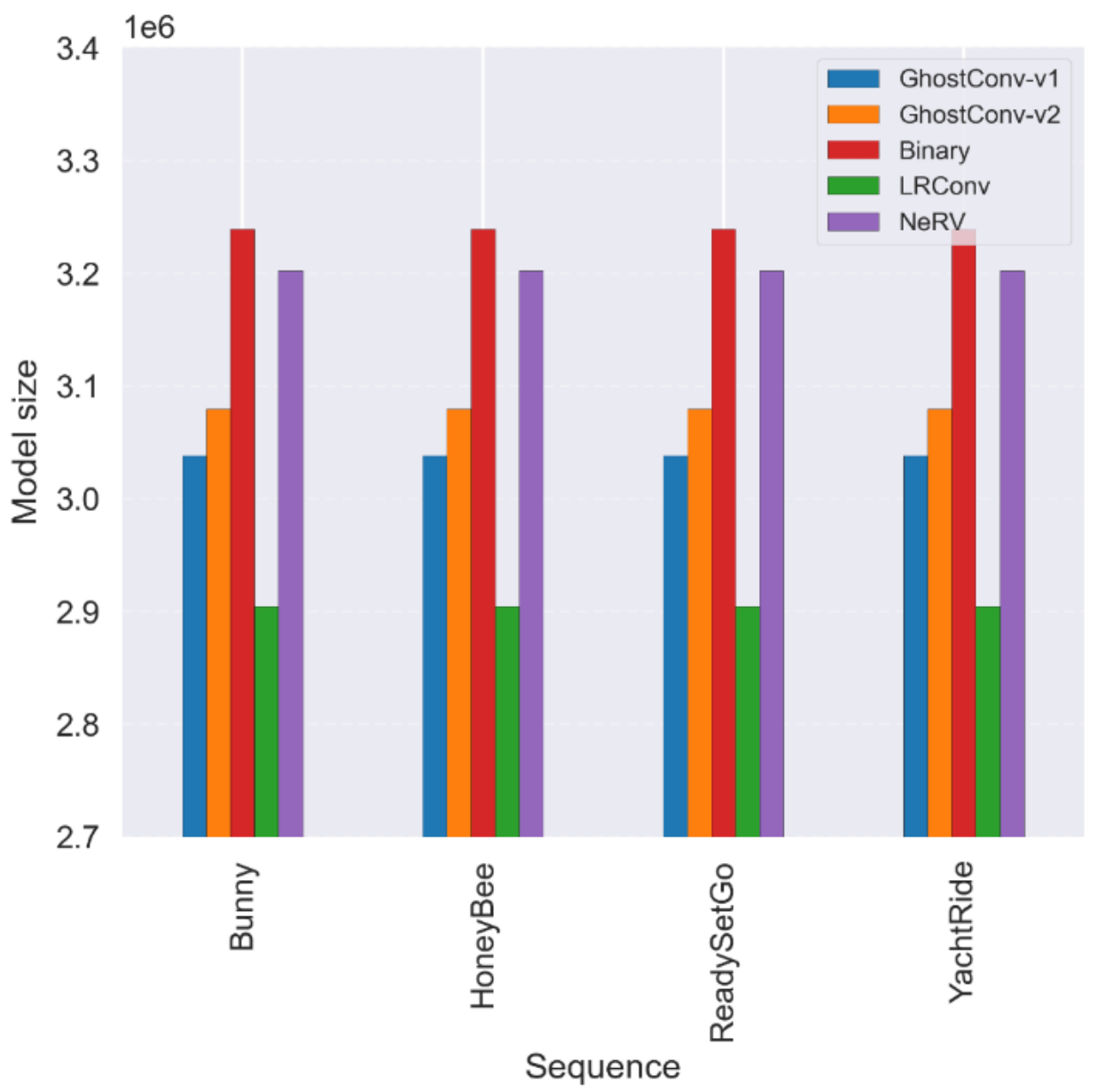

**Figure 9.** Number of parameters comparison with existing solutions applied to convolution layer 4 with INT8 quantization.

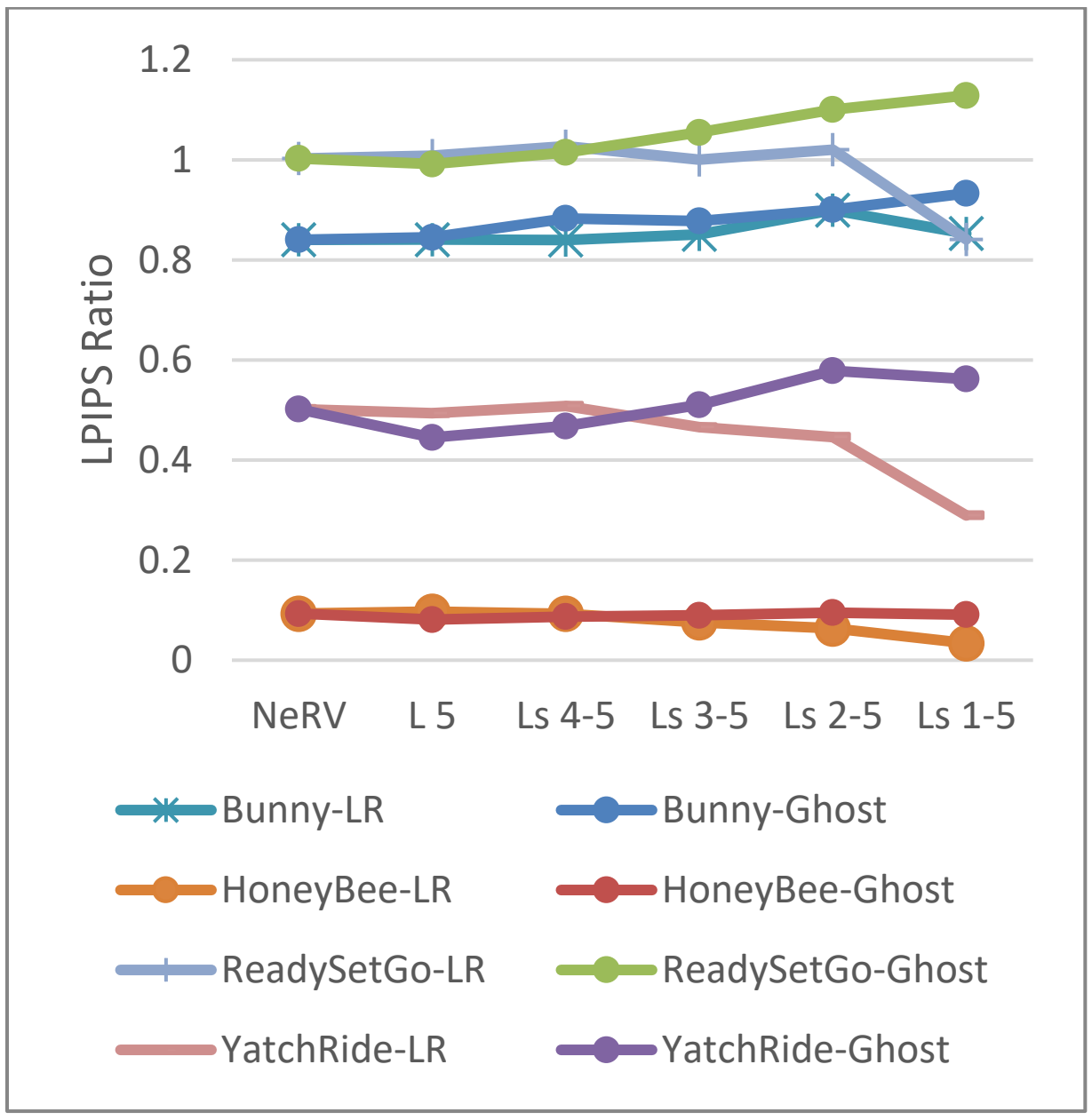

**Figure 10**. Measuring perceptual flicker across consecutive frames using the LPIPS metric

### *C. Temporal Stability*

In addition to spatial reconstruction fidelity, we evaluate the temporal stability of the reconstructed videos by measuring perceptual flicker across consecutive frames using the LPIPS metric. Specifically, we compute LPIPS between adjacent ground-truth frames $(gt_t, gt_{t-1})$ and between the corresponding reconstructed frames $(\hat{I}_t, \hat{I}_{t-1})$ using the AlexNet backbone, and report the ratio $\text{LPIPS}(\hat{I}_t, \hat{I}_{t-1})/\text{LPIPS}(gt_t, gt_{t-1})$ as a normalized measure of temporal flicker. This ratio decouples intrinsic motion in the content from reconstruction-induced temporal artifacts, enabling a fair comparison across sequences with different motion dynamics. Lower ratios indicate better temporal consistency, i.e., less flicker introduced by the reconstruction model relative to the natural temporal variation in the ground truth. We use this metric to compare the proposed LRConv–NeRV against GhostConv variants and the dense NeRV baseline, thereby assessing whether architectural factorization affects not only spatial quality but also temporal coherence in reconstructed videos.

Figure 10 shows the normalized LPIPS flicker ratio for LRConv–NeRV and Ghost-NeRV [6] compared to the NeRV baseline under progressively more aggressive layer factorization. Applying LRConv only to the largest decoder stage (L4) yields LPIPS ratios close to the baseline across sequences, indicating no noticeable increase in temporal flicker. As factorization extends to earlier stages, temporal consistency degrades in a content-dependent manner, with larger deviations observed for texture-rich and fast-motion sequences (e.g., ReadySetGo, YachtRide).

Compared to Ghost-NeRV–v1, LRConv–NeRV exhibits more stable temporal behavior under moderate factorization, while aggressive configurations introduce increased flicker. Overall, LRConv at the final decoder stage preserves temporal coherence, whereas early-stage factorization can induce temporal artifacts.

Note that Binary-NeRV [5] reports this temporal flicker metric only for the Bunny sequence, with values of 1.8, 1.5, 1.3, and 1.2 when binarization is applied to layers 1–4, 2–4, 3–4, and 4, respectively. In contrast, the corresponding values for the proposed LRConv–NeRV remain in the range of 0.85–0.9. This comparison indicates that the larger GFLOPs reduction achieved by Binary-NeRV comes at the cost of increased temporal flicker, whereas LRConv–NeRV preserves temporal coherence much closer to the dense NeRV baseline

## VI. Conclusion

This work introduced LRConv–NeRV, a low-rank convolutional variant of NeRV that reduces decoder complexity through structured architectural factorization rather than numerical approximation. By replacing selected dense convolutional layers with separable low-rank convolutions and training the resulting architecture end-to-end, LRConv–NeRV provides a controllable efficiency–quality trade-off without requiring post-hoc decomposition or specialized optimization procedures.

Experimental results on four video sequences show that applying LRConv only to the largest decoder stage achieves a strong operating point, reducing decoder complexity by approximately 68%, from 201.9 to 64.9 GFLOPs, and model size by 9.3%, while maintaining near-baseline reconstruction quality with PSNR drops below 0.5 dB and consistent bitrate reductions of about 9.2%. Under INT8 post-training quantization, LRConv–NeRV preserves reconstruction quality more robustly than aggressive early-stage factorization, which introduces disproportionate quality loss for relatively small additional savings in GFLOPs.

Compared to existing efficiency-oriented NeRV variants, LRConv–NeRV occupies a favorable middle ground. While Binary–NeRV achieves lower computational cost, it suffers from notable degradation in spatial fidelity and temporal stability. Ghost-NeRV offers competitive quality but higher GFLOPs at comparable layer configurations. In contrast, LRConv–NeRV provides substantial reductions in GFLOPs and parameters while preserving spatial detail and temporal coherence, as confirmed by LPIPS-based flicker analysis.

Overall, LRConv–NeRV offers a principled and flexible architectural mechanism for potential adaptation of NeRV decoders to resource-constrained deployment scenarios.

## ACKNOWLEDGMENT

This work is partially funded by the Open Access Program of the American University of Sharjah. This paper represents the opinions of the author does not mean to represent the position or opinions of the American University of Sharjah.

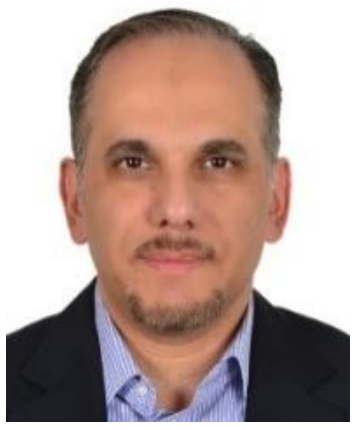

**TAMER SHANABLEH** (Senior Member, IEEE) was born in Scotland, U.K.. He received the M.Sc. degree in software engineering in 1998 and the Ph.D. degree in electronic systems engineering in 2002, both from the University of Essex, U.K.. He is currently a Professor of Computer Science with the American University of Sharjah, UAE, which he joined in 2002. From 1999 to 2002, he served as a Senior Research Officer with the University of Essex, where he collaborated with BTexact on the invention of video transcoders. He subsequently joined Motorola UK Research Labs, where he contributed to the establishment of the Error Resilient Simple Scalable Profile within the ISO/IEC MPEG-4 standard. Throughout his career, Dr. Shanableh has held several visiting professorships at Motorola Labs and completed a sabbatical leave as a visiting academic with the Multimedia and Computer Vision Lab at Queen Mary University of London, U.K.. He holds six patents and has authored more than 100 publications, including 10 IEEE Transactions papers. His primary research interests include deep learning and video coding and processing.